%% file: main.tex
\PassOptionsToPackage{table,dvipsnames}{xcolor}
\documentclass[sigconf,nonacm,pdfa]{acmart}
\usepackage{xcolor}

\newcommand{\Ecase}{{\color{BrickRed}E_{\text{case}}}}
\newcommand{\Ecode}{{\color{blue}E_{\text{code}}}}

\newcommand{\Erefinement}{{\color{YellowOrange}E_{\text{refinement}}}}

\newcommand{\Amodel}{{\color{BrickRed}A_{\text{model}}}}
\newcommand{\Arefinement}{{\color{YellowOrange}A_{\text{refinement}}}}
\newcommand{\Atune}{{\color{pink}A_{\text{fine-tune}}}}
\newcommand{\Alog}{{\color{darkgray}A_{\text{logging}}}}

\definecolor{lightred}{rgb}{0.9,0.475,0.531}
\newcommand{\ModelBox}
{\colorbox{lightred}{Model Selection}}

\newcommand{\RefinementBox}{\colorbox{Apricot}{Refinement}}
\newcommand{\TuneBox}{\colorbox{pink}{Fine-tuning}}

\definecolor{lightpurple}{rgb}{0.8, 0.6, 0.8}

\newcommand{\CodeRefineBox}{\colorbox{lightpurple}{Code Refinement}}

\usepackage{adjustbox}

\usepackage{tcolorbox}
\usepackage{listings}

\input{preamble}

\AtBeginDocument{%
  }

\begin{document}

\input{01_title}

\input{02_abstract}

\input{03_introduction}
\input{04_related_work}

\input{05_problem}

\input{06_methods}

\input{07_experiments}
\input{08_conclusions}
\input{09_acknowledgement}

\balance
\bibliographystyle{ACM-Reference-Format}
\bibliography{main}

\end{document}

%% file: preamble.tex
\usepackage{amsmath,amsfonts,amsthm}
\usepackage{graphicx}
\usepackage{textcomp}
\usepackage{balance}
\usepackage{multirow}
\usepackage{subfigure}
\usepackage{array}
\usepackage{enumitem}
\usepackage{epstopdf}
\usepackage{booktabs}
\usepackage{threeparttable}
\usepackage{bm}
\usepackage{hyperref}
\usepackage{gensymb}
\usepackage{xr}
\usepackage[capitalise]{cleveref}
\usepackage{tikz}

\usepackage{placeins}

\newtheoremstyle{mydefn}
{}{}
{\it}       
{0pt}       
{\bfseries} 
{:~}        
{0.25em}    
{}          
\theoremstyle{mydefn}

\newcounter{examplectr}
\renewcommand{\theexamplectr}{\arabic{examplectr}}

\newtcolorbox{examplebox}[2][]{%
  colback=gray!5, colframe=gray!160,
  fonttitle=\bfseries\small,
  title={Example \refstepcounter{examplectr}\label{#1}\theexamplectr: #2},
  boxrule=1pt, arc=0mm,
  left=3pt,
  right=3pt,
  top=1pt,
  bottom=1pt,
}
\makeatletter
\newif\if@restonecol
\makeatother

\usepackage[linesnumbered,ruled,vlined]{algorithm2e}
\usepackage{algpseudocode}

\SetKwComment{Comment}{$\triangleright$\ }{}
\SetKwRepeat{Do}{do}{while}

\renewcommand{\paragraph}[1]{\smallskip\noindent\textbf{#1.}}
\renewcommand{\subparagraph}[1]{\smallskip\noindent\textbf{\underline{#1.}}}

\makeatletter
\def\UrlAlphabet{%
    \do\a\do\b\do\c\do\d\do\e\do\f\do\g\do\h\do\i\do\j%
    \do\k\do\l\do\m\do\n\do\o\do\p\do\q\do\r\do\s\do\t%
    \do\u\do\v\do\w\do\x\do\y\do\z\do\A\do\B\do\C\do\D%
    \do\E\do\F\do\G\do\H\do\I\do\J\do\K\do\L\do\M\do\N%
    \do\O\do\P\do\Q\do\R\do\S\do\T\do\U\do\V\do\W\do\X%
    \do\Y\do\Z}
\def\UrlDigits{\do\1\do\2\do\3\do\4\do\5\do\6\do\7\do\8\do\9\do\0}
\g@addto@macro{\UrlBreaks}{\UrlOrds}
\g@addto@macro{\UrlBreaks}{\UrlAlphabet}
\g@addto@macro{\UrlBreaks}{\UrlDigits}
\makeatother

\usepackage[textsize=tiny]{todonotes}

\renewcommand{\paragraph}[1]{\smallskip\noindent\textbf{#1.}}
\renewcommand{\subparagraph}[1]{\smallskip\noindent\textbf{\underline{#1.}}}
\renewcommand{\sec}[1]{\textbf{\S#1}}

%% file: 01_title.tex
\title{Structured Agentic Workflows for Financial Time-Series Modeling with LLMs and Reflective Feedback}

\author{Yihao Ang}
\authornote{These authors contributed equally to this work.}
\affiliation{%
  \institution{National University of Singapore}
}
\email{yihao_ang@comp.nus.edu.sg}

\author{Yifan Bao}
\authornotemark[1]
\affiliation{%
  \institution{National University of Singapore}
}
\email{yifan_bao@comp.nus.edu.sg}

\author{Lei Jiang}
\authornotemark[1] 
\affiliation{%
  \institution{University College London}
}
\email{lei.j@ucl.ac.uk}

\author{Jiajie Tao}
\affiliation{%
  \institution{University College London}
}
\email{jiajie.tao.21@ucl.ac.uk}

\author{Anthony K. H. Tung}
\authornote{These authors are corresponding authors.}
\affiliation{%
  \institution{National University of Singapore}
}
\email{atung@comp.nus.edu.sg}

\author{Lukasz Szpruch}
\authornotemark[2]
\affiliation{%
  \institution{University of Edinburgh}
}
\email{l.szpruch@ed.ac.uk}

\author{Hao Ni}
\authornotemark[2]
\affiliation{%
  \institution{University College London}
}
\email{h.ni@ucl.ac.uk}

%% file: 02_abstract.tex
\begin{abstract}

Time-series data is central to decision-making in financial markets, yet building high-performing, interpretable, and auditable models remains a major challenge. While Automated Machine Learning (AutoML) frameworks streamline model development, they often lack adaptability and responsiveness to domain-specific needs and evolving objectives. 
Concurrently, Large Language Models (LLMs) have enabled agentic systems capable of reasoning, memory management, and dynamic code generation, offering a path toward more flexible workflow automation.
In this paper, we introduce \textsf{TS-Agent}, a modular agentic framework designed to automate and enhance time-series modeling workflows for financial applications. The agent formalizes the pipeline as a structured, iterative decision process across three stages: model selection, code refinement, and fine-tuning, guided by contextual reasoning and experimental feedback. 
Central to our architecture is a planner agent equipped with structured knowledge banks, curated libraries of models and refinement strategies, which guide exploration, while improving interpretability and reducing error propagation.
\textsf{TS-Agent} supports adaptive learning, robust debugging, and transparent auditing, key requirements for high-stakes environments such as financial services. Empirical evaluations on diverse financial forecasting and synthetic data generation tasks demonstrate that \textsf{TS-Agent} consistently outperforms state-of-the-art AutoML and agentic baselines, achieving superior accuracy, robustness, and decision traceability.
\end{abstract}

\ccsdesc[500]{Computing methodologies~Intelligent agents}
\ccsdesc[300]{Mathematics of computing~Time series analysis}

\keywords{Financial time series; LLM; Agents}

\maketitle

%% file: 03_introduction.tex
\section{Introduction}
\label{sec:introduction}

The financial industry operates at an unprecedented velocity and scale, continuously generating massive streams of time-series data. Extracting actionable and timely insights from these large, dynamically evolving datasets remains a fundamental challenge. 
Automated machine learning (AutoML) systems such as \textsf{AutoGluon}~\cite{agtimeseries} and \textsf{H2O AutoML}~\cite{H2OAutoML20} aim to reduce barriers to entry by automating feature selection and hyperparameter tuning. These systems have accelerated model development cycles and empowered non-expert users. However, their reliance on static, rule-based model selection strategies---optimized predominantly for general-purpose statistical metrics---limits their adaptability.
Recent advances in Large Language Models (LLMs) have enabled the emergence of agentic systems for data science automation. They aim to automate end-to-end workflows by coupling natural language reasoning with code generation and execution~\cite{AutoGPT, guo2024ds, baek2024researchagent, ang2024tsgassist}. For example, \textsf{ResearchAgent}~\cite{baek2024researchagent} tackles general data science workflows, while \textsf{DS-Agent}~\cite{guo2024ds} introduces a Case-Based Reasoning (CBR) paradigm that reuses historical solutions from online sources to guide LLM behavior. These frameworks reflect a growing trend toward \emph{agentic automation}, where LLMs not only interface with structured data but also orchestrate complex reasoning and decision-making pipelines.

Despite this progress, building robust, adaptive, and auditable agents for time-series modeling remains a significant open problem. In regulated environments like financial services, performance alone is insufficient. Auditable and interpretable processes are essential for ensuring regulatory compliance, enabling human-AI collaboration, and building trust. It is not only important to generate high-performing models but also to \emph{understand and justify how those models are conceived, selected, and refined}.

In this paper, we introduce \textsf{TS-Agent}, a modular agentic framework designed to automate, enhance, and audit financial time-series modeling workflows. 
Unlike traditional AutoML systems that treat modeling as a static optimization problem, our approach builds upon advances in structured reasoning systems~\cite{singh2023kaggle, guo2024ds}, extending them in two critical directions. 
First, we integrate domain-specific knowledge by incorporating curated external resources, mirroring the modularity and reuse common in real-world quantitative finance workflows. 
Second, we embed a reflective feedback mechanism, enabling the agent to iteratively update decisions based on empirical results and code execution logs, thereby improving adaptability, robustness, and transparency.

\paragraph{Key Contributions}
\textsf{TS-Agent} formalizes the workflow as a dynamic, multi-stage decision process, consisting of \emph{model selection}, \emph{code refinement}, and \emph{hyperparameter fine-tuning}, executed iteratively via a planner agent. At each stage, decisions are informed by structured reasoning, contextual memory, and performance feedback.
Our core contributions include:

\begin{itemize}[leftmargin=16pt, itemsep=2pt]
    \item \textbf{Structured Knowledge Banks for Context-Aware Decision-Making}:
    \textsf{TS-Agent} draws on three external resources to inform its decisions:  (1) a \emph{Case Bank} of past financial modeling tasks and solutions for case-based reasoning; (2) a \emph{Financial Time-Series Code Base} with executable models and metrics for direct reuse; and (3) a \emph{Refinement Knowledge Bank} encoding expert heuristics and diagnostic strategies to guide context-aware, iterative model refinement grounded in financial best practices.

    \item \textbf{Feedback-Driven Online Learning}: 
    The planner agent continually updates its policy based on feedback from experimental outcomes. This enables adaptive refinement loops that surpass the limitations of static AutoML pipelines and naive LLM-based agents, while offering a consistent interface for introspection, debugging, and improvement.
    
    \item \textbf{Auditable and Debuggable by Design}: 
    Our modular architecture isolates code modifications to specific refinement modules, while logging each decision and its rationale. This design facilitates reproducibility, fault localization, and compliance auditing---key requirements for high-stakes AI deployment in financial environments.

    \item \textbf{Empirical Validation on Financial Tasks:} 
    We evaluate \textsf{TS-Agent} across diverse real-world financial tasks---including stock price forecasting, cryptocurrency prediction, and synthetic time-series generation. Our results show that \textsf{TS-Agent} consistently outperforms AutoML pipelines and LLM-based agents in predictive accuracy, trading utility, and robustness while offering greater interpretability and success consistency.

\end{itemize}

Together, these contributions demonstrate the feasibility and advantages of agentic, interpretable, and auditable automation for time-series workflows in high-stakes domains.

%% file: 04_related_work.tex
\begin{figure*}[t]
  \centering
  \subfigure[Overall architecture of \textsf{TS-Agent}.]{%
  \label{fig:overall:architecture}%
  \includegraphics[width=0.74 \textwidth]{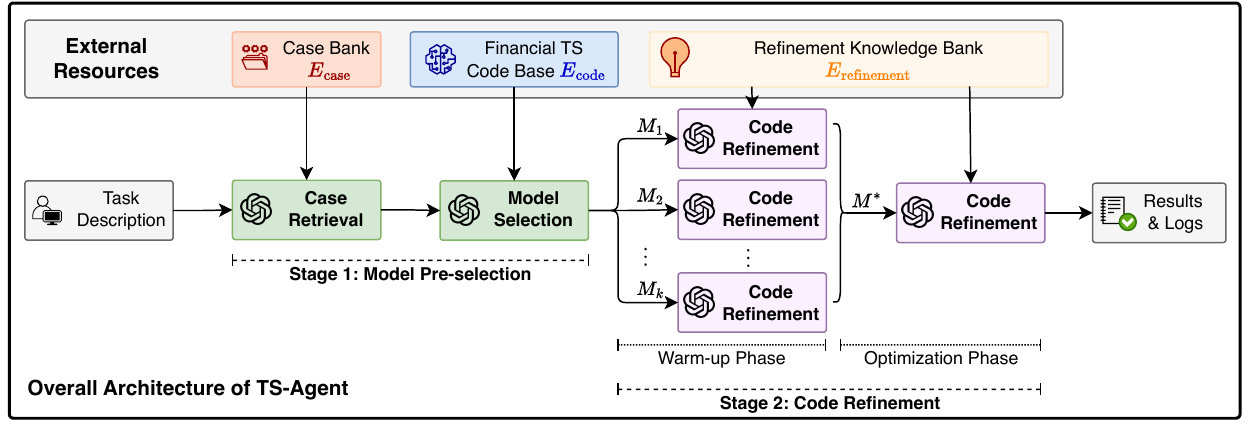}}%
  \subfigure[Feedback loop for code refinement.]{%
  \label{fig:overall:execution}%
  \includegraphics[width=0.249 \textwidth]{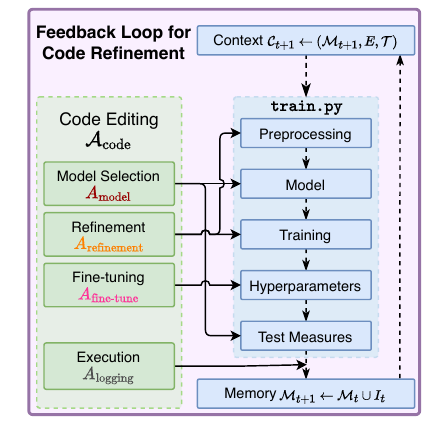}}%
  \vspace{-1.0em}
  \caption{Overview of \textsf{TS-Agent}.}
  \label{fig:overall}
  \vspace{-1.0em}
\end{figure*}

\section{Related Work}
\label{sec:related}

We review two primary lines of research related to \textsf{TS-Agent}: \textit{automated machine learning} and \textit{agentic systems}. The former focuses on automating the model development cycle, while the latter emphasizes autonomous reasoning and task execution.

\subsection{Automated Machine Learning}
\label{sec:related:automl}

Automated machine learning (AutoML) systems automate end-to-end machine learning workflows, covering data preprocessing, feature engineering, model selection, and hyperparameter optimization \cite{thornton2013auto, feurer2015efficient, H2OAutoML20, akiba2019optuna}. Typical pipelines involve (1) selecting preprocessors, (2) searching over model families, and (3) tuning hyperparameters. Recent frameworks also incorporate meta-learning and ensembling for improved generalization \cite{agtimeseries}.
Early systems such as \textsf{Auto-WEKA} \cite{thornton2013auto} and \textsf{Auto-sklearn} \cite{feurer2015efficient} formulate the model selection and hyperparameter tuning problem as a joint optimization task using Bayesian optimization. 
In time-series domains, AutoML systems typically treat forecasting as generic regression, requiring manual feature engineering for lags and trends. \textsf{AutoGluon} \cite{agtimeseries} mitigate this by integrating diverse models for probabilistic forecasting, while \textsf{Optuna}~\cite{akiba2019optuna} is an efficient and scalable hyperparameter optimization framework for generative modeling tasks.

Despite progress, existing AutoML systems face critical gaps in financial time-series tasks. 
First, they lack domain-specific reasoning capabilities. 
Second, most operate as black-box search tools with limited interpretability. 
Third, they usually optimize for statistical error metrics, often ignoring financial performance indicators.
\textsf{TS-Agent} addresses these gaps by embedding domain-specific diagnostics and financial context reasoning directly into the automation loop, moving beyond static AutoML approach, aligning model selection with financial decision-making needs.

\subsection{Agentic Systems}
\label{sec:related:agentic}

Agentic systems enable autonomous planning and execution over multi-step workflows by LLMs. Unlike static AutoML pipelines, they typically perform task decomposition, invoke external tools, and iteratively refine outputs based on feedback \cite{yao2022react, shinn2023reflexion, wang2023voyager, singh2023kaggle}.
General-purpose frameworks like \textsf{AutoGPT} \cite{AutoGPT} support autonomous goal decomposition and tool integration. The \textsf{ReAct} framework \cite{yao2022react}  interleaves reasoning traces with task-specific actions, allowing agents to dynamically plan and update their behaviour.
The \textsf{Reflexion}~\cite{shinn2023reflexion} introduced \emph{verbal reinforcement learning}, where feedback from previous episodes is transformed into natural language self-reflections, effectively encoding an interpretable memory that supports iterative learning and credit assignment. 
While general-purpose agentic systems are flexible, they often lack domain specialization.
Recent works have explored agentic systems tailored for data science and time-series tasks. For example,
\textsf{ResearchAgent} \cite{baek2024researchagent} tackled the problem of scientific ideation by constructing an LLM agent that proposes research ideas, retrieves literature, and iteratively refines its outputs through self-critique and multi-agent peer review. 
\textsf{DS-Agent} \cite{guo2024ds} introduced a case-based reasoning framework for automated data science. It retrieves relevant cases from a human insight database, adapts them to new tasks, and iteratively revises the solution based on execution feedback.

Current agentic systems, however, are not tailored for high stakes decision making. The  often lack transparency, adaptability, and effective human–AI teaming, making it ill-suited for deployment in high-stakes financial services. Furthermore, for these system to be adapted then need to integrate with already existing code bases and align with organizations specific know-hows.

%% file: 05_problem.tex
\section{Problem formulation}
\label{sec:problem}
The main objective of the \textsf{TS-Agent} is to automatically solve any given financial time-series tasks accurately and effectively by leveraging the external library information (including the contextual information and code base) and the flexible agent workflow. 

\subsection{Financial Time-Series Tasks}
A machine learning task typically consists of a task description ($\mathcal{T}$), the dataset ($\mathcal{D}$), and an evaluation criterion for model performance ($\mathcal{L}$).
Here, the dataset $\mathcal{D}:= (\mathcal{D}_{\text{train}}, \mathcal{D}_{\text{test}})$ contains the information about the training and test split. Model training is performed exclusively on the training dataset, and once training is complete, the evaluation criterion $\mathcal{L}$ is used to assess the model's performance on the test dataset. 
In this paper, we primarily focus on two categories of financial time-series tasks, described below. The evaluation criteria may include general-purpose metrics for each task type or be tailored to specific financial applications, depending on the use case. 
We offer concrete examples of evaluation measures in \sec{\ref{sec:exp}}.

\paragraph{Time-Series Forecasting Tasks} 
The forecasting task refers to the prediction of the future time-series given the past ones. Let $X:= (X_t)_{t \in \mathbb{N}}$ denote a $d$-dimensional time-series. It is the regression task where the input and output pairs are $(X_{t-p: t}, X_{t+1, t+q})_{t \in \mathfrak{T}}$, where $\mathfrak{T}$ is the set of available time stamps. The objective of this task is to predict the conditional expectation of $\mathbb{E}[X_{t+1:t+q} | X_{t-p+1:t}]$ given the past time-series information $X_{t-p:t}$ as accurately as possible. Let $F_{\theta}: \mathbb{R}^{d \times p} \rightarrow \mathbb{R}^{d \times q}$. 
One commonly used evaluation metric is Mean Squared Error ($\text{MSE}$), which is defined as 
\begin{displaymath}
\text{MSE}:=\frac{1}{\mathfrak{T}} \sum_{t \in \mathfrak{T}}||F_{\theta}(X_{t-p+1:t}) - X_{t+1, t+q}||^2.
\end{displaymath}
In financial applications, the additional task-specific metrics (e.g., profit-and-loss measures or risk-adjusted scores) are often used.

\paragraph{Time-Series Generation Tasks}
It aims to construct a generative model that produces synthetic time-series whose distribution closely matches that of the true time-series based on the observations. 
Suppose that we aim to learn the distribution of $X_{\text{seg}}:=X_{t:t+q}$, which is assumed to be independent of $t$. The dataset in this case consists of samples $(X_{t:t+q})_{t \in \mathfrak{T}}$. The generative model is composed of the pre-specified noise distribution $\xi \in \mathcal{P}(E)$ and the generator, denoted by $G_{\theta}$, is a function mapping from $E \rightarrow \mathbb{R}^{d \times q}$. 
Then, it induces the fake distribution $G_{\theta}(\xi)$. The task is to select an appropriate noise distribution and design and train a generator $G_{\theta}$ so that $G_{\theta}(\xi)$ produces high-fidelity samples of $X_{t: t+q}$. 
Evaluation metrics for this task may include statistical distance measures (e.g., Wasserstein distance, maximum mean discrepancy) or task-specific financial criteria.

\subsection{Learning Framework}
In practice, financial institutions typically maintain their own code pipelines for analyzing financial time-series data and have their domain knowledge bases. To reflect it, in our work, we assume that the agent has access to three read-only external resources:
\begin{enumerate}[leftmargin=16pt, itemsep=2pt]
    \item \textbf{Case Bank} (denoted by $\Ecase$, textual data) as a collection of past tasks and reports summarizing successful methodologies.
    
    \item \textbf{Refinement Knowledge Bank} (denoted by $\Erefinement$, textual data) as common practices in preprocessing, training optimization, and hyperparameter tuning and evaluation.

    \item \textbf{Code Base} (denoted by $\Ecode$), a repository containing model implementations of various models (model bank) and evaluation metrics (evaluation measures bank) (in \texttt{.py} format).
    
\end{enumerate}

\begin{figure*}[t]
  \centering
  \subfigure[Case Bank.]{%
  \label{fig:bank:case}%
  \includegraphics[width=0.2226 \textwidth]{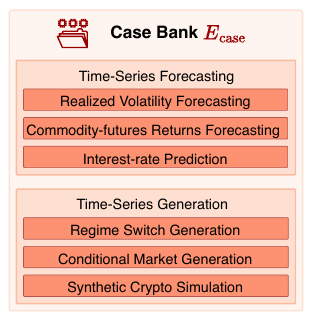}}%
  \subfigure[Financial Time-Series Code Base.]{%
  \label{fig:bank:codebase}%
  \includegraphics[width=0.475 \textwidth]{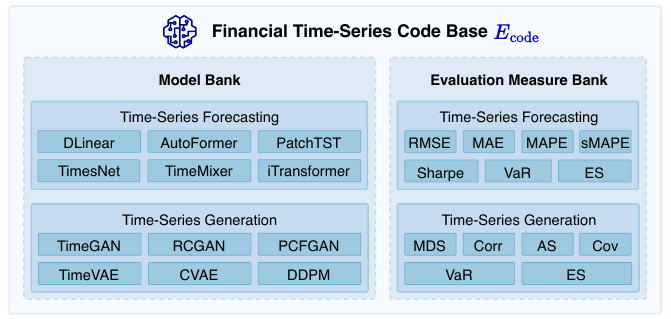}}%
  \subfigure[Refinement Knowledge Bank.]{%
  \label{fig:bank:refinement}%
  \includegraphics[width=0.2898 \textwidth]{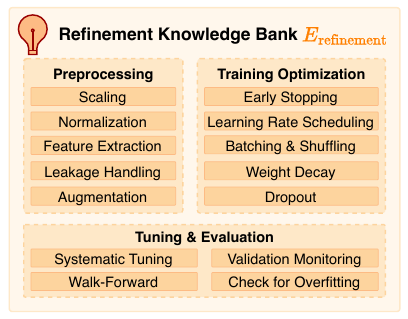}}%
  \vspace{-1.25em}
  \caption{Key components in the three external resources.}
  \label{fig:bank}
  \vspace{-1.25em}
\end{figure*}

Given a new task $T$ (see Example \ref{ex:task} as illustration) and at $t=0$, the agent is given a Python template $\text{train.py}$ (denoted by $\mathcal{S}_0$, see Example \ref{ex:train_py} as illustration) along with the above external resources. Using this information, the agent's objective is to complete the submodules of $\mathcal{S}_0$ so that it becomes executable, successfully trains the model, and produces valid model predictions with the goal of minimizing the corresponding loss function on the test dataset. 
For clarity, we use color coding to highlight information associated with key actions, which we defer to define in \sec{\ref{sec:methods}}. Below, we will provide the details of each external source. An illustrative overview of these external sources is provided in Figure \ref{fig:bank}.

\paragraph{Case Bank}
\label{sec:problem:knowledge}
Similar to existing agentic systems \cite{guo2024ds}, \textsf{TS-Agent} possesses a \emph{Case Bank} $\mathcal{C}=\{c_1,\ldots,c_{|\mathcal{C}|}\}$ as a library consisting of the financial time-series tasks and the reports of successful methodologies. These cases could be derived from industry benchmarks, top-performing competition entries, and peer-reviewed academic papers on relevant financial data analysis (e.g., \cite{ni2023regime, ni2024market, ni2024crypto, Ang2023TSGBench}). 
It comprises two types of cases: time-series forecasting and time-series generation. Its tasks involve diverse financial datasets, such as historical stock prices, exchange rates, cryptocurrencies, and synthetic data. For instance, one task in our case bank is a task of generative time-series modelling on the cryptocurrency markets, where the recommended solution is a VAE-based model. Figure~\ref{fig:bank:case} shows an illustration of the case bank.

\paragraph{Refinement Knowledge Bank} 
A library of best practices and tips for improving the training of models. As illustrated in Figure \ref{fig:bank:refinement}, it contains guidance on selecting preprocessing methods, training strategies, optimization techniques, and evaluation metrics based on the logged training results. 
For example, a guidance entry under Learning Rate Scheduling might be:
\textit{Employ a learning rate scheduler (e.g., \texttt{ReduceLROnPlateau}) to decrease the learning rate when validation performance plateaus. This facilitates smoother convergence.} The refinement knowledge bank is divided into three components:
\begin{itemize}[leftmargin=16pt, itemsep=2pt]
    \item \textbf{Preprocessing}: Scaling, normalization for financial indicators, and data augmentation to address heavy-tailed returns, microstructure noise, and sparsity in high-frequency data.

    \item \textbf{Training Optimization}: Early stopping based on losses, adaptive learning rates for regime shifts, and regularization (e.g., dropout, weight decay) to mitigate overfitting.

    \item \textbf{Tuning and Evaluation}: Cross validation, hyperparameter tuning, and overfitting diagnostics to ensure robust out-of-sample forecasting and trading performance.
\end{itemize}

\paragraph{Code Base}
The code base includes the implementation of a model bank and an evaluation measure bank. 

\begin{itemize}[leftmargin=16pt, itemsep=2pt]

    \item \textbf{Model Bank} is a curated collection of implemented financial models.
    It includes the most representative models of time-series forecasting and generative models. For the forecasting task, there could be several main types, including the tree-based models, deep learning based models, and econometric models. For the generative models, the typical models contain GAN-based, VAE-based, and diffusion-based ones. The model bank is designed to ensure broad coverage of state-of-the-art techniques for financial time-series analysis.
    
    \item \textbf{Evaluation Measure Bank} offers a diverse set of evaluation metrics, allowing users to select those most appropriate for different financial use cases.  It includes commonly used statistical metrics and financial-specific test metrics. For example, for the generative task, it includes risk-aware tail metrics, temporal dependency measures, and distributional similarity metrics.

\end{itemize}

%% file: 06_methods.tex
\section{\textsf{TS-Agent} Framework}
\label{sec:methods}
The \textsf{TS-Agent} consists of two stages: (1) the model pre-selection stage, and (2) the code refinement stage, depicted in Figure \ref{fig:overall:architecture}. 
The agent iteratively updates $\texttt{train.py}$ by leveraging both external resources and the training performance from previous executions of $\texttt{train.py}$. The final output of the \textsf{TS-Agent} consists of reproducible code for generating high-performing models along with auditable logs that record the development process, actions, and underlying reasoning.
To enhance the agent's effectiveness, we exploit the structured design of $\texttt{train.py}$ and implement a chain-of-code-edits mechanism that systematically guides the refinement process, analogous to the chain of thought.

\paragraph{Action Space} 
The action space of the agent includes the following four main actions:

\begin{itemize}[leftmargin=16pt, itemsep=2pt]
\item $\Amodel$: select the models and measures in $\texttt{train.py}$;
\item $\Arefinement$: select training tips in $\texttt{train.py}$;
\item $\Atune$: select the hyper-parameter configurations;       
\item $\Alog$: execute $\texttt{train.py}$ and log the code development and the training results of the experiment.
\end{itemize}

As the $\texttt{train.py}$ is modular code template, the action of code editing $\mathcal{A}_{\text{code}}$ can be decomposed into the above first three submodules $\Amodel$, $\Atune$ and $\Arefinement$. 
Only $\Arefinement$ may introduce the coding bugs, which increases the controllability and reliability of our agents. 
In this case, the agent will debug and fix, i.e., report the error and generate the fix iteratively, following the approach in \citet{guo2024ds}. Moreover, $\Alog$ includes recording the code changes and the reason for the change and the corresponding train results, which would be useful for auditing purposes.

\paragraph{Dynamic Memory and Context}
At each iteration $t$, the agent conducts the action of code editing. 
Once the $\texttt{train.py}$ is executable, the agent proceeds with $\Alog$ and the training log of each experiment (e.g., various test metrics along with the relational for the code changes) denoted by $I_{t}$, is recorded. 
Recall that $\mathcal{S}_t$ denote the $\texttt{train.py}$ at time $t$. The memory of the agent at time $t$ is defined as $\mathcal{M}_t:=(I_v, \mathcal{S}_v)_{v \in t}$, which dynamically incorporates the new experiment results, the codes/configuration of each experiment and all the log information generated by the agents. 

The \emph{context} at time $t$, denoted by $\mathcal{C}_t$ is defined as the union of the dynamic memory information $\mathcal{M}_t$ and the static external library $E$, task information $\mathcal{T}$. This context $\mathcal{C}_t$ provides the agent with the valuable information to iteratively update $\texttt{train.py}$.

\paragraph{Feedback Loop and Chain-of-code-edits}
Let $\pi(A | \mathcal{M}_t, E)$ denote the conditional probability of the agent to conduct the action $A$ based on the information $I$. At time $t$, the conditional probability of the code edits given the context information satisfies that
\begin{eqnarray*}
\pi(\mathcal{A}_{\text{code}} | \mathcal{C}_t)  
=& \pi(\Amodel | \mathcal{C}_t) \cdot \pi(\Atune | \Amodel, \mathcal{C}_t )  \\
&\cdot \pi(\Arefinement | \Amodel, \Atune, \mathcal{C}_t).
\end{eqnarray*}

Due to the high dimensionality of the contextual information $\mathcal{C}_t$, we approximate each term on the right-hand side using the parsimonious conditioning information, which affects the action. For each term, we carefully select the most relevant subset of $\mathcal{C}_t$ and remove redundant or negligible contextual information, thereby effectively reducing dimensionality while preserving utility.
Based on this formulation, we illustrate the feedback loop in Figure~\ref{fig:overall:execution}, comprising three key modules:

\begin{enumerate}
    \item \ModelBox. For efficient filtering, the key information for selecting promising model candidates is the past case library $\Ecase$ and the task description $\mathcal{T}$. We employ case-based reasoning to enable the agent to recommend the most suitable models ($\Amodel$).

    \item \RefinementBox. This step
    involves modifying the code related to refinement ($\Arefinement$) using knowledge from the refinement tip library $\Erefinement$ and the additional the training log of numerical experiments $I_t$.

   \item \TuneBox. The agent recommends a hyper-parameter configuration based on the training results in $I_t$, which include multiple test metrics, and updates the hyper-parameter configuration file (\texttt{.yaml}) through $\Atune$. 
\end{enumerate}
\sloppy Once the \textsf{TS-Agent} completes \TuneBox, it executes the code and records the training results ($\Alog$). Then, the agent's log information $I_t$ is incorporated into its memory. The pseudocode of this feedback loop is summarized in Algorithm~\ref{Algo_feedback}.
\ModelBox is conducted at Stage 1, while \RefinementBox and \TuneBox form the key building block (i.e., Code Refinement) of Stage 2. The details of each module are elaborated in \sec{\ref{sec:methods:stage1}} and \sec{\ref{sec:methods:stage2}}, respectively.

\begin{algorithm}
\SetAlgoLined
\caption{Feedback Loop for \CodeRefineBox.}\label{Algo_feedback}
Initialize contextual information $\mathcal{C}_1$\;
\For{$t \gets 1$ \KwTo $T_{\text{max}}$}{
    Conduct \ModelBox, \RefinementBox, \TuneBox\;
    Conduct action $\Alog$ and record the log $I_t$\;
    Update the agent's memory $\mathcal{M}_{t+1} \gets \mathcal{M}_t  \cup I_t$\;
    Update the context $\mathcal{C}_{t+1} \gets (\mathcal{M}_{t+1}, E, \mathcal{T})$\;
}
\end{algorithm}
\vspace{-1.0em}

\subsection{Stage 1: Model Pre-selection}
\label{sec:methods:stage1}

At Stage 1, the agent performs \ModelBox and applies \emph{case-based reasoning} \cite{guo2024ds} to recommend the top-$k$ model candidates for the given task $\mathcal{T}$. By leveraging existing well-established financial models from the case bank $\Ecase$, it retrieves similar past problems and efficiently narrows down the most promising options in the model pool of the code base. The agent then specifies each suggested model $M_i$ separately in $\texttt{train.py}$ for $i \in \{1, \cdots, k\}$.

\subsection{Stage 2: Code Refinement}
\label{sec:methods:stage2}

Stage 2 employs a two-phase Round-Robin Search \cite{rasmussen2008round}, consisting of the Warm-up Phase and the Optimization Phase.
Its core building block is \CodeRefineBox, which iteratively applies \RefinementBox and \TuneBox. If a candidate script reduces the loss, it replaces the incumbent and serves as the starting point for the next refinement-and-fine-tuning cycle; otherwise, the change is rejected and the search reverts to the previous script. An illustrative case study is provided in \sec{\ref{sec:exp:case}} and Figure~\ref{fig:stage2_example}. 

\paragraph{Warm-up Phase}
The agent first conducts \CodeRefineBox \emph{in parallel} on each model $M_i$ for a small number of iterations. The best combination of model, refinement, and hyper-parameters is then selected based on the evaluation criteria of the given task.  

\paragraph{Optimization Phase}
In this stage, the agent continues to perform \CodeRefineBox but for a larger number of iterations, which is crucial for further enhancing the training of the chosen model to achieve high accuracy.

%% file: 07_experiments.tex
\input{tables/tsf}

\input{tables/tsf_crypto}

\section{Experiments}
\label{sec:exp}

We conduct extensive experiments to evaluate the effectiveness, robustness, and versatility of \textsf{TS-Agent} in financial time-series analytics, compared to AutoML and state-of-the-art agentic systems.
In addition, we present a case study showcasing the auditable logs of the agent's model selection, refinement, and tuning process, highlighting the added transparency and interpretability benefits of \textsf{TS-Agent} beyond its high predictive performance.

\subsection{Experimental Setup}
\label{sec:exp:setup}

\paragraph{Task Selection}
We evaluate \textsf{TS-Agent} across a diverse set of financial time-series tasks, which fall into two main categories: forecasting and generation. The forecasting task assesses the agent's ability to produce accurate, risk-aware predictions in forward-looking market scenarios. It is critical for applications such as asset pricing, volatility forecasting, and trading signal generation \cite{Hu2025FinTSB, wang2025fintsbridge}.
The generation task focuses on simulating synthetic financial time-series that preserve realistic statistical and structural characteristics. It is particularly valuable for stress testing, risk simulation, and data augmentation under scarce or proprietary conditions \cite{Ang2023TSGBench, sig-gan}.

\paragraph{Datasets}
We employ three financial datasets from diverse sources:
\begin{itemize}[leftmargin=16pt, itemsep=2pt]

    \item \textbf{Crypto} \cite{ctbench}: Hourly closing prices for a set of 20 cryptocurrency trading pairs against USDT (Tether) in 2024. 
    
    \item \textbf{Exchange} \cite{lai2018modeling}: Daily foreign exchange rates from 1990 to 2010 for eight major currency pairs, each quoted against U.S. Dollar.
    
    \item \textbf{Stock} \cite{timegan}: Daily closing prices for 10 major U.S. stocks from January 2020 to December 2024, covering business days only.
    
\end{itemize}

\paragraph{Baselines and LLM Backbones}
To evaluate the performance of \textsf{TS-Agent}, we benchmark it against two representative agentic systems: \textbf{\textsf{DS-Agent}}~\cite{guo2024ds}, and \textbf{\textsf{ResearchAgent}}~\cite{baek2024researchagent}.
In addition, we include two widely adopted AutoML baselines: \textbf{\textsf{AutoGluon}}~\cite{agtimeseries} for time-series forecasting, and \textbf{\textsf{Optuna}}~\cite{akiba2019optuna} for time-series generation. 

All agents are instantiated using four popular large language models: \textbf{GPT-3.5} and \textbf{GPT-4o} from OpenAI, \textbf{Claude Sonnet 4} from Anthropic, and \textbf{Nova Pro} from Amazon.

\input{tables/tsg_regime_stock}

\subsection{Time Series Forecasting}
\label{sec:exp:tsf}

\paragraph{Model Bank}
We include five representative forecasting models that span the major paradigms in time-series forecasting: two transformer-based models (\textbf{Autoformer} \cite{wu2021autoformer} and \textbf{PatchTST} \cite{patchtst}), a temporal CNN backbone (\textbf{TimesNet} \cite{wutimesnet}), a linear model (\textbf{DLinear} \cite{dlinear}), and a token-mixing model (\textbf{TimeMixer} \cite{wangtimemixer}).

\paragraph{Evaluation Measures}
To comprehensively assess \textsf{TS-Agent}'s forecasting performance, we adopt four metrics, \textbf{RMSE}, \textbf{MAE}, \textbf{MAPE}, and \textbf{sMAPE}, commonly used to evaluate prediction accuracy \cite{wangtimemixer}, and report their averages over all assets and test windows.
For Crypto dataset, we additionally assess financial fidelity via risk/trading metrics computed on prediction-implied returns: \textbf{Sharpe}, \textbf{Value at Risk (VaR)}, and \textbf{Expected Shortfall (ES)} \cite{ni2024crypto}. We present differences of these metrics between prediction and ground truth.
Lastly, we report the \textbf{Success Rate} over 5 runs for each dataset.

\begin{figure}[t]
  \centering
  \includegraphics[width=0.75\columnwidth]{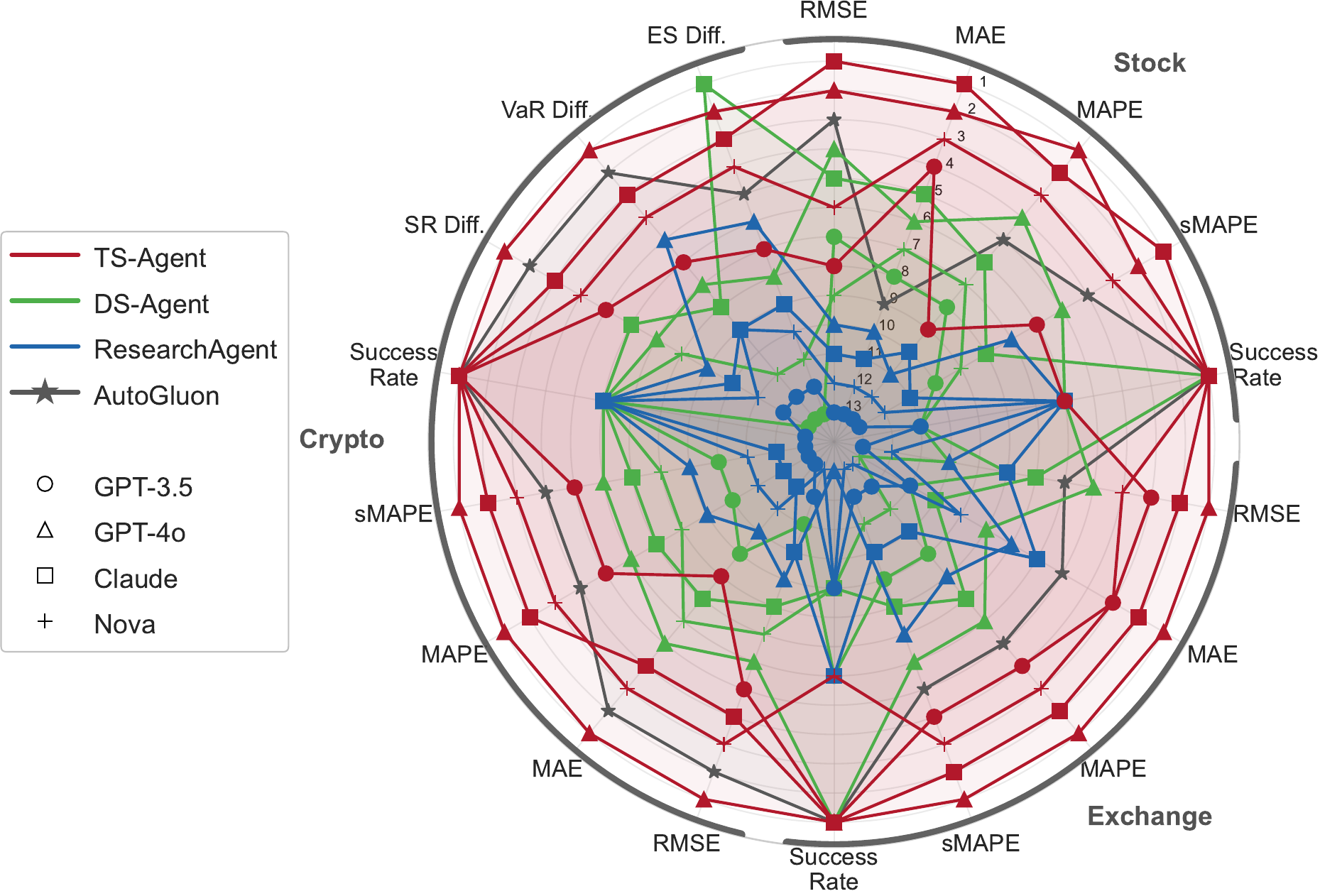}
  \vspace{-0.5em}
  \caption{Rankings of agents and \textsf{AutoGluon} across datasets and measures on time-series forecasting tasks.}
\label{fig:tsf_radar}
  \vspace{-0.75em}
\end{figure}

\paragraph{Overall Results}
Table~\ref{tab:tsf_all} demonstrates that \textsf{TSAgent} outperforms every baseline across three datasets with its best LLM variant (usually GPT-4o or Claude) while maintaining an 100\% success rate over five runs. Compared with \textsf{AutoGluon}, \textsf{TSAgent} cuts RMSE by more than 20\% on Exchange and around 8\% on Crypto. Versus competitor agents, it reduce RMSE by up to 30\% over \textsf{DS-Agent} and 15–40\% over \textsf{ResearchAgent}. This combination of superior accuracy and stability underscores the effectiveness of \textsf{TSAgent} for multivariate time-series forecasting.
These improvements underscore the effectiveness of structured agentic workflows in aligning model selection and refinement with domain-specific objectives.
Table~\ref{tab:tsf_trading} further highlights the robustness of \textsf{TS-Agent} in preserving risk-sensitive structures, particularly in volatile environments such as cryptocurrency markets. With GPT-4o, \textsf{TS-Agent} achieves the lowest Sharpe Ratio Difference and VaR Difference, both around 20\% better than the strongest competing agent, while maintaining a competitive ES. 
These consistent gains across both predictive and risk metrics indicate that \textsf{TS-Agent} not only produces more accurate forecasts but also better retains critical properties essential for downstream financial decision-making.

\paragraph{Ranking Analysis}
Figure~\ref{fig:tsf_radar} illustrates the ranking profiles across all metrics and datasets. \textsf{TS-Agent} forms the outermost contour on most axes, securing the best overall average rank. \textsf{DS-Agent} trails by 1–3 positions but remains competitive with \textsf{AutoGluon}, while \textsf{ResearchAgent} consistently clusters near the center.
Among LLM backbones, GPT-4o yields the top performance across agents, though the margin between GPT-4o and other LLMs is narrow for \textsf{TS-Agent} and more pronounced for \textsf{DS-Agent} and \textsf{ResearchAgent}. 
We attribute this backbone-agnostic resilience to the inclusion of the Financial TS Code Base, which lets \textsf{TS-Agent} refine existing models rather than creating them from scratch, thus reducing variance.

\input{tables/tsg_crypto}

\subsection{Time Series Generation}
\label{sec:exp:tsg}

\paragraph{Model Bank} We include five representative generative models covering the major paradigms in the field: three GAN-based models (\textbf{TimeGAN} \cite{yoon2019time}, \textbf{PCFGAN} \cite{lou2023pcf}, and \textbf{RCGAN} \cite{esteban2017real}), a VAE-based model (\textbf{TimeVAE} \cite{desai2021timevae}), and a diffusion-based model (\textbf{DDPM} \cite{ho2020denoising}).

\paragraph{Evaluation Measures}
We evaluate the fidelity of synthetic time-series on the Exchange and Stock datasets using four statistical metrics: \textbf{Marginal Distribution Score}, \textbf{Correlation Score}, \textbf{Autocorrelation Score}, and \textbf{Covariance Score}, which measure the $\ell_2$ differences between real and generated time-series in marginal distributions, feature-wise autocorrelations, temporal covariances, and cross-feature correlations, respectively \cite{sig-gan, timegan}. For the Crypto dataset, we further evaluate the fitting of tail behavior using \textbf{VaR} and \textbf{ES}, due to the heavy-tailed nature of cryptocurrency returns.

\paragraph{Overall Results} 
Tables~\ref{tab:tsg_exchange_stock} and \ref{tab:tsg_crypto} demonstrate that the \textsf{TS-Agent} outperforms other agents across nearly all datasets, LLM types and evaluation metrics in terms of generation quality. 
Compared to the efficient hyperparameter tuning framework \textsf{Optuna}, \textsf{TS-Agent} shows comparable or even superior performance, highlighting the effectiveness of \textsf{TS-Agent} in selecting both an appropriate model and optimal hyperparameters.
\textsf{TS-Agent} also establishes strong robustness in success rate across different LLMs, consistently achieving a 100\% success rate, while the success rates of \textsf{DS-Agent} and \textsf{Research-Agent} vary widely from 0\% to 100\%. Although all agents reach 100\% success with Claude, \textsf{TS-Agent} still achieves significantly better performance: it outperforms competitors by 77.39\% and 83.7\% on the Correlation metric for the Stock dataset, and by 55.99\% and 16.90\% on the Exchange dataset. 
Moreover, \textsf{TS-Agent} proves more stable performance across different LLMs. For instance, its Correlation score on the Stock dataset ranges from 1.194 to 3.468, compared to $[5.282, 11.991]$ for \textsf{DS-Agent} and $[7.175, 12.879]$ for \textsf{ResearchAgent}. This reduced variance stems from \textsf{TS-Agent}'s structured design, which leverages a model bank and modular code template to minimize code generation from scratch. 
These results confirm that \textsf{TS-Agent} not only achieves superior generation quality but also maintains strong robustness and consistency across diverse LLMs and datasets, making it a reliable and fully automated solution for time-series generation in financial markets.

\begin{figure}[t]
  \centering
  \includegraphics[width=0.75\columnwidth]{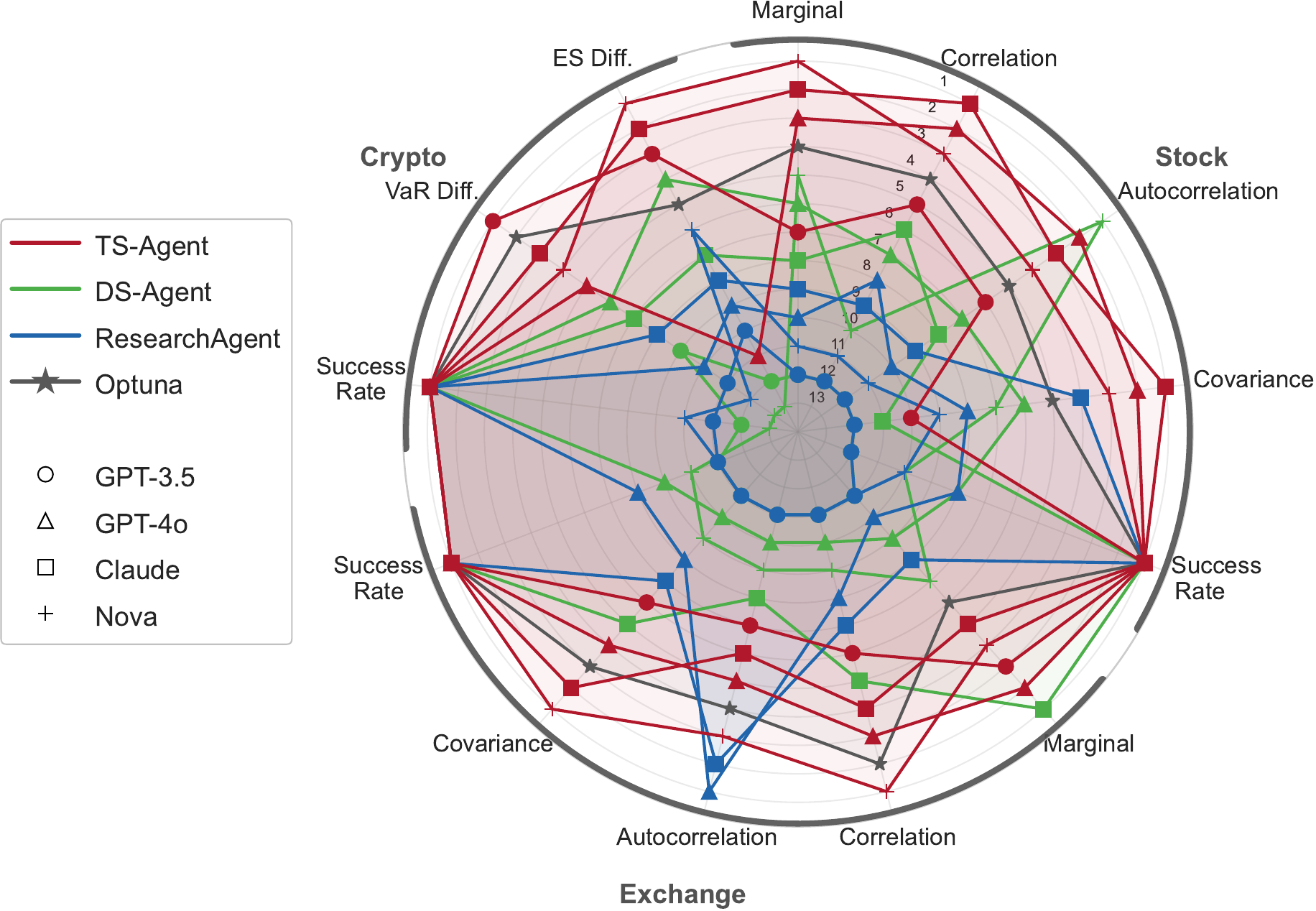}
  \vspace{-0.75em}
  \caption{Rankings of agents and \textsf{Optuna} across datasets and measures on time-series generation tasks.}
\label{fig:tsg_radar}
  \vspace{-1.0em}
\end{figure}

\begin{figure*}[t]
  \centering
  \includegraphics[width=0.99\textwidth]{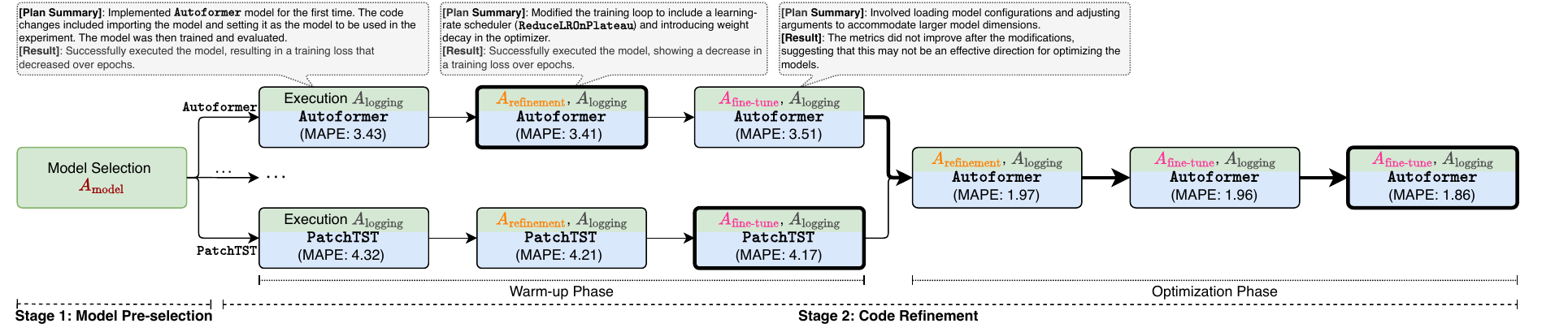}
  \caption{An illustration of \textsf{TS-Agent}'s two-stage workflow on a financial time-series forecasting task.}
\label{fig:stage2_example}
\end{figure*}

\paragraph{Ranking Analysis} 
Figure~\ref{fig:tsg_radar} summarizes the rankings of agents and the \textsf{Optuna} baseline across datasets and metrics for time-series generation tasks. 
\textsf{TS-Agent} again dominates with the outermost contour across almost all spokes, confirming its consistent top rank. \textsf{Optuna} ranks the second-best overall, with a wider contour than both \textsf{DS-Agent} and \textsf{ResearchAgent} across many metrics—particularly on Stock and Crypto datasets. \textsf{DS-Agent} is positioned third overall, performing competitively on some axes but generally ranking lower than \textsf{Optuna}, especially in terms of coverage and correlation.
These results suggest that domain-informed agentic workflows, as implemented in \textsf{TS-Agent}, offer greater generalization and robustness than the AutoML or generic agentic baselines.

\input{case_study/case_study_1_task}

\input{case_study/case_study_1_train_py}

\subsection{Case Study}
\label{sec:exp:case}
We now conduct a case study on a financial time-series forecasting task (Example~\ref{ex:task}) that predicts the next three trading-day closes of ten U.S. stocks using a 60-day input window. Performance is evaluated using average MAPE.
\textsf{TS-Agent} starts from a scaffolded pipeline (Example~\ref{ex:train_py}) with placeholder definitions for dataset, model, and training loop.
As shown in Figure~\ref{fig:stage2_example}, Stage 1 performs model pre-selection via case-based reasoning using the task description. \texttt{Autoformer} and \texttt{PatchTST} are shortlisted. The choice is informed by prior cases with similar task structure and model suitability.

In Stage 2, the agent runs a two-phase round-robin procedure, alternating between model refinements and fine-tuning, each followed by validation and feedback logging.
During the Warm-up Phase, each shortlisted model undergoes the feedback loop consisting of refinement, fine-tuning, and execution. Each step is logged to track performance. The logs (i.e., gray boxes in Figure~\ref{fig:stage2_example}) illustrate how the agent learns from both performance gains and non-beneficial changes, enabling reflective learning, adaptive planning, and effective performance optimization.
Only the best-performing version of each candidate (e.g., \texttt{Autoformer} at MAPE 3.41, \texttt{PatchTST} at MAPE 4.17) is retained and compared. The superior model (\texttt{Autoformer}) is then selected for the Optimization Phase, where it undergoes additional iterative refinements.
This phase further reduces MAPE to 1.86, as the agent filters out ineffective changes and builds upon beneficial edits.

This case highlights how \textsf{TS-Agent}'s modular design and structured decision-making enable efficient and interpretable end-to-end automation while preserving interpretability and auditability in financial applications.

%% file: tables/tsf.tex
\begin{table*}[t]
  \centering
  \caption{Time series forecasting performance on three benchmark datasets. Each metric is averaged over five runs. The best result for each LLM (per column) across different agents is bolded.}
  \vspace{-0.75em}
  \resizebox{\textwidth}{!}{%
    \begin{tabular}{cccccccccccccccccccccc}
    \toprule
    \multirow{2}[3]{*}{\textbf{Dataset}} & \multirow{2}[3]{*}{\textbf{Model}} & \multicolumn{4}{c}{\textbf{RMSE} $\downarrow$}       & \multicolumn{4}{c}{\textbf{MAE} $\downarrow$}      & \multicolumn{4}{c}{\textbf{MAPE} (\%) $\downarrow$}       & \multicolumn{4}{c}{\textbf{sMAPE} (\%) $\downarrow$}      & \multicolumn{4}{c}{\textbf{Success Rate} (\%) $\uparrow$} \\
\cmidrule(lr){3-6}\cmidrule(lr){7-10}\cmidrule(lr){11-14}\cmidrule(lr){15-18}\cmidrule(lr){19-22}          &       & GPT-3.5 & GPT-4o & Claude & Nova & GPT-3.5 & GPT-4o & Claude & Nova & GPT-3.5 & GPT-4o & Claude & Nova & GPT-3.5 & GPT-4o & Claude & Nova & GPT-3.5 & GPT-4o & Claude & Nova \\
    \midrule
    \multirow{3}[2]{*}{Crypto} & \textsf{TS-Agent} & \textbf{0.277} & \textbf{0.206} & \textbf{0.254} & \textbf{0.249} & \textbf{0.0076} & \textbf{0.0051} & \textbf{0.0066} & \textbf{0.0063} & \textbf{1.692} & \textbf{1.531} & \textbf{1.643} & \textbf{1.655} & \textbf{1.693} & \textbf{1.529} & \textbf{1.644} & \textbf{1.658} & \textbf{100} & \textbf{100} & \textbf{100} & \textbf{100} \\
          & \textsf{DS-Agent} & 0.369 & 0.297 & 0.320 & 0.307 & 0.0082 & 0.0070 & 0.0074 & 0.0073 & 2.065 & 1.883 & 1.905 & 1.912 & 2.063 
& 1.881 & 1.907 & 1.914 & 60 & 60 & 60 & 60 \\
          & \textsf{ResearchAgent} & 0.392 & 0.341 & 0.355 & 3.477 & 0.0094 & 0.0083 & 0.0088 & 0.0085 & 2.344 & 2.040 & 2.210 & 2.198 & 
2.347 & 2.040 & 2.230 & 2.200 & 40 & 60 & 60 & 60 \\
          & \textsf{AutoGluon} & \multicolumn{4}{c}{0.223}        & \multicolumn{4}{c}{0.0055}        & \multicolumn{4}{c}{1.664}        & \multicolumn{4}{c}{1.662}        & \multicolumn{4}{c}{100} \\
    \midrule
    \multirow{3}[2]{*}{Exchange} & \textsf{TS-Agent} & \textbf{0.0073} & \textbf{0.0068} & \textbf{0.0069} & \textbf{0.0077} & \textbf{0.0041} & \textbf{0.0036} & \textbf{0.0040} & \textbf{0.0041} & \textbf{0.564} & \textbf{0.474} & \textbf{0.499} & \textbf{0.562} & \textbf{0.563} & \textbf{0.474} & \textbf{0.498} & \textbf{0.560} & \textbf{100} & \textbf{100} & \textbf{100} & \textbf{80} \\
          & \textsf{DS-Agent} & 0.0095 & 0.0088 & 0.0090 & 0.0096 & 0.0066 & 0.0057 & 0.0062 & 0.0068 & 0.801 & 0.782 & 0.792 & 0.811 & 0.800 & 0.781 & 0.792 & 0.810 & 80 & \textbf{100} & 60 & 60 \\
          & \textsf{ResearchAgent} & 0.0099 & 0.0096 & 0.0095 & 0.0098 & 0.0066 & 0.0056 & 0.0054 & 0.0059 & 0.812 & 0.793 & 0.804 & 0.817 & 0.811 & 0.79 & 0.804 & 0.817 & 60 & 40 & 80 & 60 \\
          & \textsf{AutoGluon} & \multicolumn{4}{c}{0.0089}        & \multicolumn{4}{c}{0.0052}        & \multicolumn{4}{c}{0.701}        & \multicolumn{4}{c}{0.699}        & \multicolumn{4}{c}{100} \\
    \midrule
    \multirow{3}[2]{*}{Stock} & \textsf{TS-Agent} & \textbf{8.727} & \textbf{8.017} & \textbf{7.982} & \textbf{8.590} & \textbf{5.117} & \textbf{4.912} & \textbf{4.905} & \textbf{5.047} & 2.336 & \textbf{2.046} & \textbf{2.076} & \textbf{2.123} & \textbf{2.001} & \textbf{1.770} & \textbf{1.765} & \textbf{1.850} & \textbf{80} & \textbf{100} & \textbf{100} & \textbf{100} \\
          & \textsf{DS-Agent} & 8.644 & 8.557 & 8.559 & 8.732 & 5.248 & 5.193 & 5.150 & 5.207 & \textbf{2.308} & 2.137 & 2.177 & 2.244 & 2.142 & 1.969 & 2.055 & 2.099 & 60 & 80 & \textbf{100} & 60 \\
          & \textsf{ResearchAgent} &  9.890 & 9.410 & 9.570 & 9.791 & 6.041 & 5.677 & 5.738 & 5.910 & 2.878 & 2.498 & 2.420 & 2.590 & 2.414 & 
2.053 & 2.238 & 2.331 & 60 & 80 & 80 & 80 \\
          & \textsf{AutoGluon} & \multicolumn{4}{c}{8.430}        & \multicolumn{4}{c}{5.258}        & \multicolumn{4}{c}{2.174}        & \multicolumn{4}{c}{1.890}        & \multicolumn{4}{c}{100} \\
    \bottomrule
    \end{tabular}%
    }
    \vspace{-0.75em}
  \label{tab:tsf_all}%
\end{table*}%

%% file: tables/tsf_crypto.tex
\begin{table}[t]
  \centering
  \caption{Trading performance on time series forecasting task for Crypto dataset.}
  \vspace{-0.75em}
  \resizebox{0.99\columnwidth}{!}{%
    \begin{tabular}{cccccccccccccccccc}
    \toprule
    \multirow{2}[3]{*}{\textbf{Dataset}} & \multirow{2}[3]{*}{\textbf{Model}} &  \multicolumn{4}{c}{\textbf{Sharpe Ratio Difference} $\downarrow$} & \multicolumn{4}{c}{\textbf{VaR Difference} $\downarrow$}       & \multicolumn{4}{c}{\textbf{ES Difference} $\downarrow$} \\
\cmidrule(lr){3-6}\cmidrule(lr){7-10}\cmidrule(lr){11-14}\cmidrule(lr){15-18}          &       & GPT-3.5 & GPT-4o & Claude & Nova & GPT-3.5 & GPT-4o & Claude & Nova & GPT-3.5 & GPT-4o & Claude & Nova \\
    \midrule
    \multirow{3}[1]{*}{Crypto} & TS-Agent    & \textbf{0.414}    & \textbf{0.378}    & \textbf{0.403}    & \textbf{0.407}    & \textbf{0.0088}    & \textbf{0.0069}    & \textbf{0.0077}    & \textbf{0.0080}    & \textbf{0.0082}    & \textbf{0.0066}    & 0.0067    & \textbf{0.0069} \\
          & DS-Agent  & 0.484    & 0.46    & 0.458    & 0.462    & 0.0130    & 0.0089   & 0.0092    & 0.0098    & 0.0121    & 0.0084    & \textbf{0.0061}   & 0.0089 \\
          & ResearchAgent   & 0.479     & 0.471    & 0.472    & 0.475    & 0.0117    & 0.0086    & 0.0094    & 0.0090    & 0.0111    & 0.0081    & 0.0087    & 0.0088 \\
          & AutoGluon &  \multicolumn{4}{c}{0.402}        & \multicolumn{4}{c}{0.0076}        & \multicolumn{4}{c}{0.0072} \\
    \bottomrule
    \end{tabular}%
    }
  \label{tab:tsf_trading}%
  \vspace{-0.75em}
\end{table}%

%% file: tables/tsg_regime_stock.tex
\begin{table*}[t]
  \centering
  \caption{Time series generation performance on Exchange and Stock datasets. \textemdash{} indicates that no successful runs were recorded. Each metric is averaged over five runs. The best result for each LLM (per column) across different agents is bolded.}
  \vspace{-0.75em}
  \label{tab:tsg_exchange_stock}
  \resizebox{\textwidth}{!}{%
    \begin{tabular}{cccccccccccccccccccccc}
    \toprule
    \multirow{2}[3]{*}{\textbf{Dataset}} & \multirow{2}[3]{*}{\textbf{Model}} 
    & \multicolumn{4}{c}{\textbf{Marginal} $\downarrow$}  
    & \multicolumn{4}{c}{\textbf{Correlation} $\downarrow$} 
    & \multicolumn{4}{c}{\textbf{Autocorrelation} $\downarrow$} 
    & \multicolumn{4}{c}{\textbf{Covariance} $\downarrow$} 
    & \multicolumn{4}{c}{\textbf{Success Rate} (\%) $\uparrow$} \\
    \cmidrule(lr){3-6} \cmidrule(lr){7-10} \cmidrule(lr){11-14} \cmidrule(lr){15-18} \cmidrule(lr){19-22}
     &  & GPT-3.5 & GPT-4o & Claude & Nova 
        & GPT-3.5 & GPT-4o & Claude & Nova 
        & GPT-3.5 & GPT-4o & Claude & Nova 
        & GPT-3.5 & GPT-4o & Claude & Nova 
        & GPT-3.5 & GPT-4o & Claude & Nova \\
    \midrule
    \multirow{4}{*}{Exchange} 
        & \textsf{TS-Agent}       &  \textbf{0.360}   &  \textbf{0.356}   &  0.374   & \textbf{0.366}    &  \textbf{4.249}   &  \textbf{2.796}   &  \textbf{2.803}   &  \textbf{1.874}   &  \textbf{0.00426}   &  0.00216   &  0.00226   &  \textbf{0.00120}   &  \textbf{0.00301}   &   \textbf{0.00179}  &  \textbf{0.00151}   &   \textbf{0.00120}  &   \textbf{100}  &   \textbf{100}  &  \textbf{100}   &  \textbf{100}   \\
        & \textsf{DS-Agent}       &  \textemdash{}   &  0.692   &  \textbf{0.329}   &  0.401  &  \textemdash{}   &  14.755   &  3.373   & 14.290    &  \textemdash{}   &  0.0605    &  0.00475   &  0.0347   &  \textemdash{}   &  0.00706  &  0.00287   &  0.00600   &  0   &   40  &  \textbf{100}   &  20   \\
        & \textsf{ResearchAgent}  &  \textemdash{}   &    0.737 &  0.530   &  \textemdash{}   &  \textemdash{}   &  9.591   &  6.369   &  \textemdash{}   &  \textemdash{}   &   \textbf{0.000884}  &  \textbf{0.00105}   &  \textemdash{}   &  \textemdash{}   &   0.00595  &  0.00403   &  \textemdash{}   &  0   &  80   &  \textbf{100}   &  0   \\
        & \textsf{Optuna} & \multicolumn{4}{c}{0.377}        & \multicolumn{4}{c}{2.114}        & \multicolumn{4}{c}{0.00198}        & \multicolumn{4}{c}{0.00172}        & \multicolumn{4}{c}{100} \\
    \midrule
    \multirow{4}{*}{Stock} 
        & \textsf{TS-Agent}       &   \textbf{0.309}  & \textbf{0.269}    &  \textbf{0.266}   &   \textbf{0.259} &   \textbf{3.468}  &  \textbf{1.228}   &  \textbf{1.194}   &  \textbf{1.413}   &  \textbf{0.227}   &  \textbf{0.152}   &  \textbf{0.153}   &  0.161   &  $\bm{6.99 \times 10^{-5}}$    &  $\bm{4.43\times 10^{-5}}$   &  $\bm{4.35\times 10^{-5}}$   &  $\bm{4.50\times 10^{-5}}$   &  \textbf{100}   &  \textbf{100}   &   \textbf{100}  &  \textbf{100}   \\
        & \textsf{DS-Agent}       &  \textemdash{}   &  0.299   &  0.331   &   0.277  &   \textemdash{}  &  6.930   &  5.282   &  11.991   &  \textemdash{}   &  0.231   &  0.286   &  \textbf{0.147}   &  \textemdash{}   &   5.58$\times 10^{-5}$  &  5.81$\times 10^{-4}$   &   5.63$\times 10^{-5}$  &  0   &  80   &   \textbf{100}  &  40   \\
        & \textsf{ResearchAgent}  &  \textemdash{}   & 0.575    &  0.397   &  0.904   &   \textemdash{}  &  7.175   &  7.343   & 12.879    &  \textemdash{}   &  0.490  &  0.329  &  1.244    &  \textemdash{}   &   5.75$\times 10^{-5}$   &  4.74$\times 10^{-5}$   &  5.84$\times 10^{-5}$   &  0   &  80   &   \textbf{100}  &  40   \\
        & \textsf{Optuna} & \multicolumn{4}{c}{0.270}        & \multicolumn{4}{c}{1.530}        & \multicolumn{4}{c}{0.162}        & \multicolumn{4}{c}{$4.83\times10^{-5}$}        & \multicolumn{4}{c}{100} \\
    \bottomrule
    \end{tabular}%
  }
  \vspace{-0.75em}
\end{table*}

%% file: tables/tsg_crypto.tex
\begin{table}[t]
  \centering
  \caption{Time series generation performance on Crypto dataset. \textemdash{} indicates that no successful runs were recorded.}
  \vspace{-0.5em}
  \resizebox{0.99\columnwidth}{!}{%
    \begin{tabular}{ccccccccccccc}
    \toprule
    \multirow{2}[3]{*}{\textbf{Model}}  & \multicolumn{4}{c}{\textbf{VaR Difference} $\downarrow$}       & \multicolumn{4}{c}{\textbf{ES Difference} $\downarrow$}        & \multicolumn{4}{c}{\textbf{Success Rate} (\%) $\uparrow$} \\
\cmidrule(lr){2-5}\cmidrule(lr){6-9}\cmidrule(lr){10-13}         & GPT-3.5 & GPT-4o & Claude & Nova & GPT-3.5 & GPT-4o & Claude & Nova & GPT-3.5 & GPT-4o & Claude & Nova\\
    \midrule
    \textsf{TS-Agent} & \textbf{0.00205}    &   \textbf{0.00326}    &  \textbf{0.00235}    &  \textbf{0.00245}   &  \textbf{0.000714}   &  0.00299   &   \textbf{0.000629}    &   \textbf{0.000566}    &  \textbf{100}     &   \textbf{100}   &  \textbf{100}  &  \textbf{100}   \\
    \textsf{DS-Agent} &  0.00564   &  0.00378     &  0.00427    &   \textemdash{}  &  0.00617   &   \textbf{0.000729}  &  0.00115     &    \textemdash{}   &   20    &   \textbf{100}   &  \textbf{100}  &  0    \\
    \textsf{ResearchAgent} &  0.0107   &   0.0105    &   0.00557   &  0.0126   &   0.00227  &   0.00149  &   0.00147    &   0.00113    &   40    &   \textbf{100}   &  \textbf{100}  &  60 \\
    \textsf{Optuna} & \multicolumn{4}{c}{0.00220}        & \multicolumn{4}{c}{0.000832}         & \multicolumn{4}{c}{100} \\
    \bottomrule
    \end{tabular}%
    }
  \label{tab:tsg_crypto}%
  \vspace{-0.5em}
\end{table}%

%% file: case_study/case_study_1_task.tex
\begin{examplebox}[ex:task]{Task Description.}
{\scriptsize
\begin{lstlisting}[breaklines,basicstyle=\ttfamily,]
You are requested to design a deep learning model to forecast the closing prices of U.S. stock for the next 3 trading days using the past 60 business days of data from 10 major companies. 
The preprocessed dataset provides input - target pairs with shapes [N, 60, 10] for inputs and [N, 3, 10] for targets.
Evaluate your model's performance using average MAPE across all forecasted days and companies.
Complete the scaffolded pipeline in train.py by implementing or refining the dataset loader, model, and training loops.
\end{lstlisting}
}
\end{examplebox}

%% file: case_study/case_study_1_train_py.tex
\begin{examplebox}[ex:train_py]{The provided Python script (\texttt{train.py}).}
{\scriptsize
\begin{lstlisting}[breaklines,basicstyle=\ttfamily,language=Python,]
# Import necessary packages
# --- Dataset Definition ---
class TimeSeriesForecastingDataset(Dataset):
    def __init__(self, csv_path, seq_len, pred_len, ...):
        # 1. Load and preprocess data
        # 2. Split data by train/val/test flag
        # 3. Normalize if needed
        # 4. Generate time features if needed
# --- DataLoader Builder ---
def build_data_loader(args):
    # Create train, val, test datasets and wrap in DataLoader
    return train_loader, val_loader, test_loader
# --- Main Training Function ---
def run(algo):
    args = load_config(algo)
    # --- Training Loop ---
    for epoch in range(args.epochs):
        # Train for one epoch
        for batch in train_loader:
            # Forward, compute loss, backward, optimize
        # Validation
        with torch.no_grad():
            for batch in val_loader:
                # Evaluate loss and store predictions
        # Save best model
    # --- Test Evaluation ---
    load_best_model()
    test_pipeline(args, model, test_loader, ...)
if __name__ == '__main__':
    run("MODEL_NAME")
\end{lstlisting}
}
\end{examplebox}

%% file: 08_conclusions.tex
\section{Conclusion and Future Work}
\label{sec:conclusions}

This paper introduces \textsf{TS-Agent}, a modular, structured agentic framework designed to automate financial time-series workflows. By formalizing the modeling pipeline as a multi-stage decision process, \textsf{TS-Agent} integrates domain-specific knowledge and feedback-driven reasoning to deliver interpretable, adaptive, and high-performing solutions across forecasting and generation tasks. 
Looking ahead, we plan to extend \textsf{TS-Agent} to support more task families such as fraud detection, integrate multimodal financial data, and explore automatic code synthesis by mining research papers and public repositories to enrich the model and refinement banks.

%% file: 09_acknowledgement.tex
\begin{acks}
HN and LJ are supported by the EPSRC [grant number EP/S026347/1]. HN is also supported by The Alan Turing Institute under the EPSRC grant EP/N510129/1. 
Moreover, HN and LJ gratefully acknowledge the support of the Impact Accelerator program at the UCL Centre for Digital Innovation, powered by Amazon Web Services. 
This project is partially supported by the AWS computing resources. They especially thank Tomasz Grzybowski (AWS) and Igor Tseyzer (UCL) for their valuable insights on the implementation of the LLM agent workflow. Besides, HN and LJ thank Hang Lou for useful discussions on the evaluation bank design.  
In addition, AT, YA, and YB are supported by the Ministry of Education, Singapore, under its MOE AcRF TIER 1 Grant (T1 251RES2517). Any opinions, findings and conclusions or recommendations expressed in this material are those of the author(s) and do not reflect the views of the Ministry of Education, Singapore. 
Moreover, LS acknowledges the support of the UKRI Prosperity Partnership Schemes: EP/V056883/1 - FAIR Framework for responsible adoption of Artificial Intelligence in the financial services industry and  APP43592: AI² – Assurance and Insurance for Artificial Intelligence, which supported this work.
\end{acks}